\pdfoutput=1
\documentclass{article}

\usepackage[preprint]{nips_2018}

\usepackage[utf8]{inputenc} % allow utf-8 input
\usepackage[T1]{fontenc}    % use 8-bit T1 fonts
\usepackage{hyperref}       % hyperlinks
\usepackage{url}            % simple URL typesetting
\usepackage{booktabs}       % professional-quality tables
\usepackage{amsfonts}       % blackboard math symbols
\usepackage{nicefrac}       % compact symbols for 1/2, etc.
\usepackage{microtype}      % microtypography
\usepackage{graphicx}
\title{Automatic Inspection of Utility Scale Solar Power Plants using Deep Learning}

\author{
  Alekh Karkada Ashok \\
  \texttt{alekhka@gmail.com} \\
  \And
  Chandan G \\
  \texttt{chandan.g78@gmail.com } \\
  \And
  Adithya Bhat \\
  \texttt{aditya@14weeks.in} \\
  \And
  Kausthubh Karnataki\\
  \texttt{kaustubh@14weeks.in} \\
  \And
  Ganesh Shankar \\
  \texttt{sganesh@alum.iisc.ac.in} \\
}
\begin{document}

\maketitle

\begin{abstract}
  Solar energy has the potential to become the backbone energy source for the world. Utility scale solar power plants (more than 50 MW) could have more than 100K individual solar modules and be spread over more than 200 acres of land. Traditionally methods of monitoring each module becomes too costly in the utility scale. We demonstrate an alternative using the recent advances in deep learning to automatically analyze drone footage. We show that this can be a quick and reliable alternative. We show that it can save huge amounts of power and the impact the developing world hugely.
\end{abstract}

\section{Introduction}

Solar energy is a clean, safe and renewable alternative to fossil fuels. It is not a surprise that it is being increasingly used to power the world. It is expected to cover 27\% of world's energy needs by 2050 \cite{frankl2010technology}. Such an energy source can become the backbone of many developing countries and is unsurprisingly one of the fastest developing industries in India. India has solar power plants of 21GW capacity and has five power plants of the top 10 largest power plants in the world.

With 60GW of the 100GW planned solar target by 2022 coming from utility scale solar PV plants and the tariff falling significantly, it’s very important to model different challenges faced in such huge plants and find scalable solutions to address these challenges. Cost optimization in project commissioning and operation and maintenance(O\&M) becomes a critical part of the plan and different technologies have been proposed over time to solve these challenges in a scalable way.

In this work we show how deep learning based object detection can yield great results in detection of defects in solar panels and how it is suitable to be applied in various parts of the world to improve the energy output of solar power plants.

\section{Defects in solar panels}

Utility scale power plants are spread across hundreds of acres of land and typically have hundreds of thousands of modules. It is very important to inspect these modules to identify the defects they have developed. Typical modules are rated for 1\% degradation per year while actual degradation can be higher. These defects accumulate over time and if left unchecked can mean serious loss in energy output.

Traditionally solar modules were monitored by attaching devices to every module but in the utility scale this becomes too expensive. An alternative to monitor by bunching many modules together which has disadvantage of being inaccurate. Since these defects show up as hotspots (they are at a different temperature compared to other parts of the module) in thermal camera imagery, we can use them to identify the location of these defects. Figure 1 shows some defects and their thermal signature.
\begin{figure}
  \centering
  
  \includegraphics[width=0.7\paperwidth]{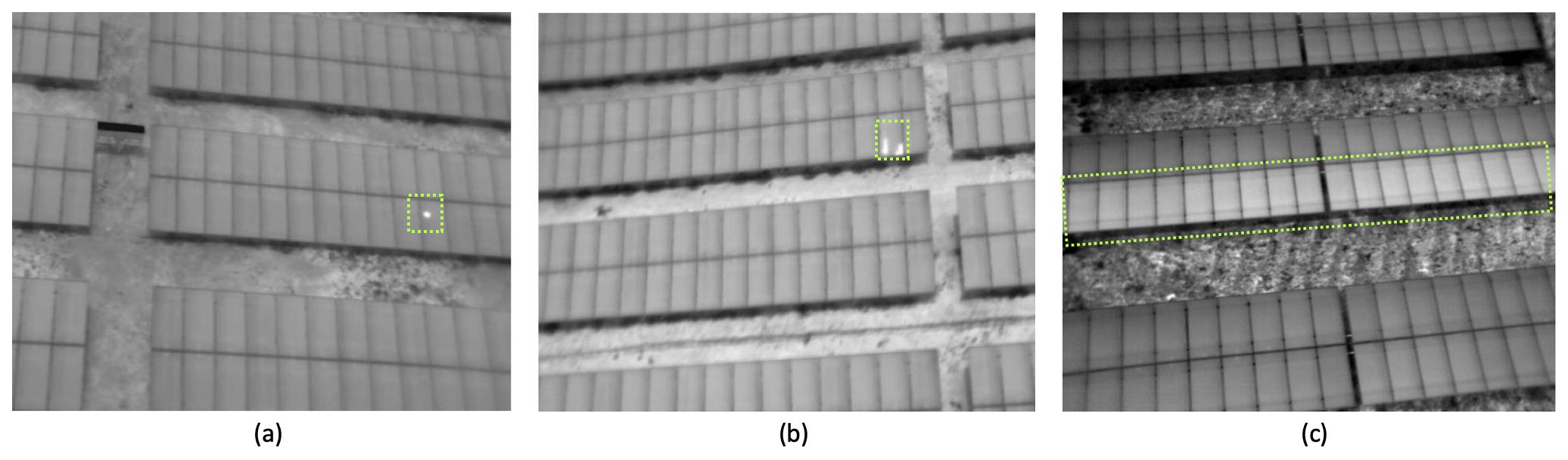}
  \caption{Typical thermal images of defects: (a) Single cell hotspot (b) Multi cell hotspot (c) Module hotspot}
  \label{typ_defects}
\end{figure}

\section{Implementation}
Object detection aims to both localize and classify objects in a given image. There are several methods to predict bounding box around objects and the class they belong to. Recent advances in deep learning \cite{Szegedy2013DeepNN, DBLP:journals/corr/SermanetEZMFL13, Erhan2014ScalableOD, Szegedy2014ScalableHO} has improved object detection and opened up many use case where it can be applied. There are also a plethora of methods to choose from, each suitable for different use cases. Methods like Faster RCNN \cite{ren2015faster} are highly accurate while but have relatively large prediction time while methods like YOLO \cite{redmon2016you} sacrifice accuracy for speed.

\subsection{Dataset}

Dataset was collected using a thermal camera attached to a custom built drone. Various solar power plants around India were used to collect the data. 16 bit TIFF images obtained from the thermal camera were contrast stretched and converted to 8 bit JPEG images. Over 2000 images per class were annotated for 3 classes: single cell hotspot, multi cell hotspot and module hotspot.

\subsection{Training}

We chose YOLO model owing to it's speed. Pretrained Convolutional layers which were used for the network. The whole network was trained using 5500 images and 500 were used for testing. Further, few image augmentation techniques are used on the training set to improve the robustness of the model.

\section{Results}

The model gives satisfactory and consistent results for a variety of input images. We observe that the model has converges quickly when YOLO anchors are calculated to the training set.

\begin{figure}
  \centering
  
  \includegraphics[width=0.6\paperwidth]{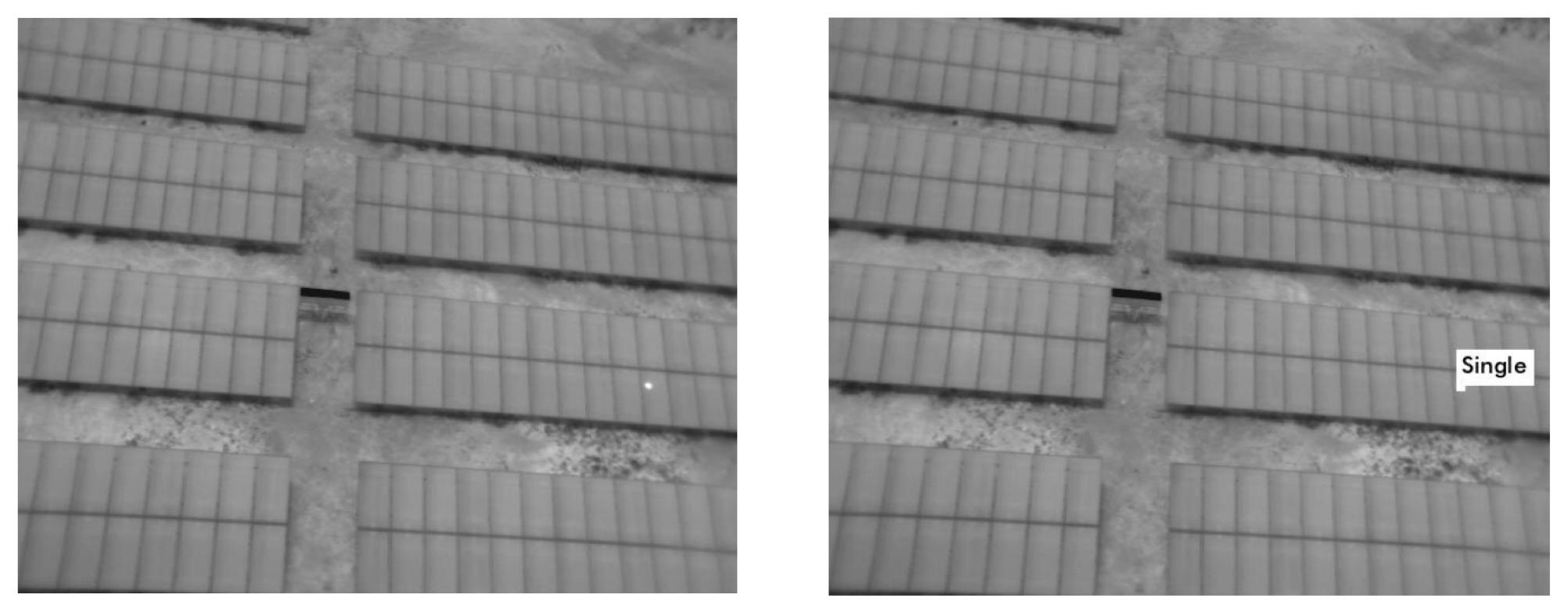}
  \includegraphics[width=0.6\paperwidth]{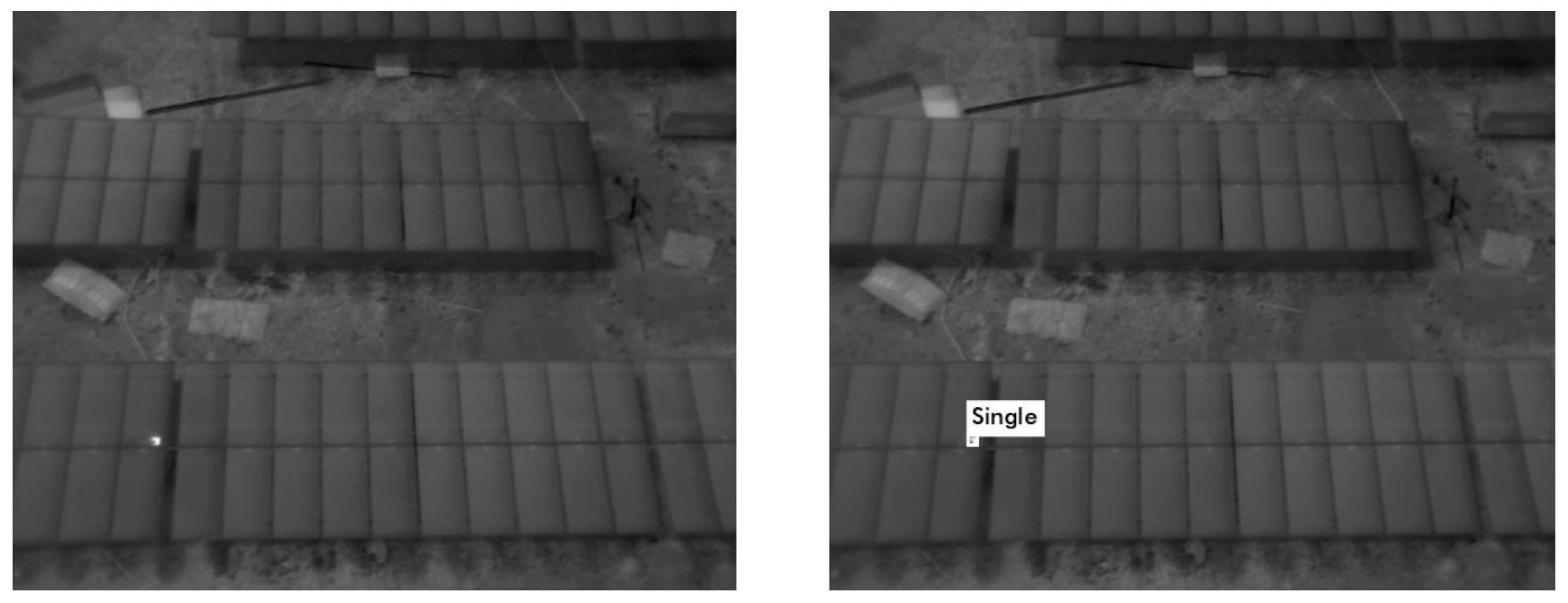}
  \caption{Model outputs for single cell hotspots.}
  \label{styp_defects}
\end{figure}

\begin{figure}
  \centering
  
  \includegraphics[width=0.6\paperwidth]{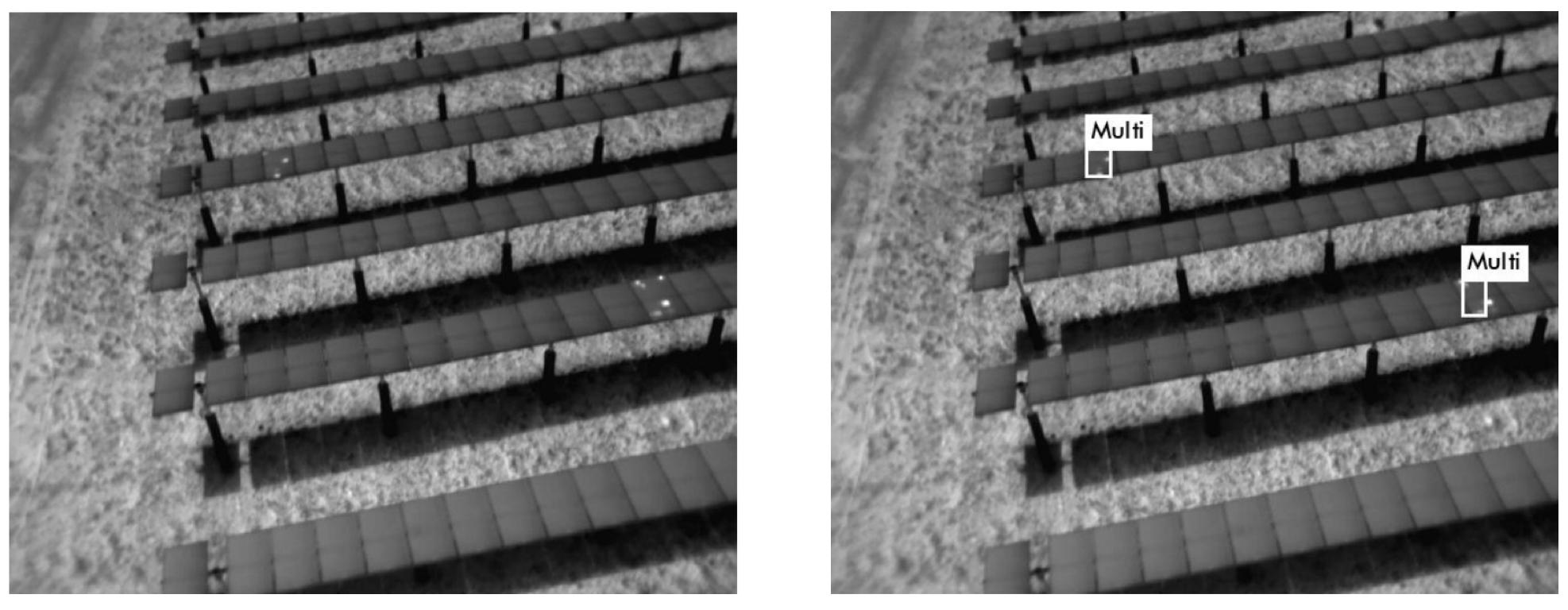}
  \includegraphics[width=0.6\paperwidth]{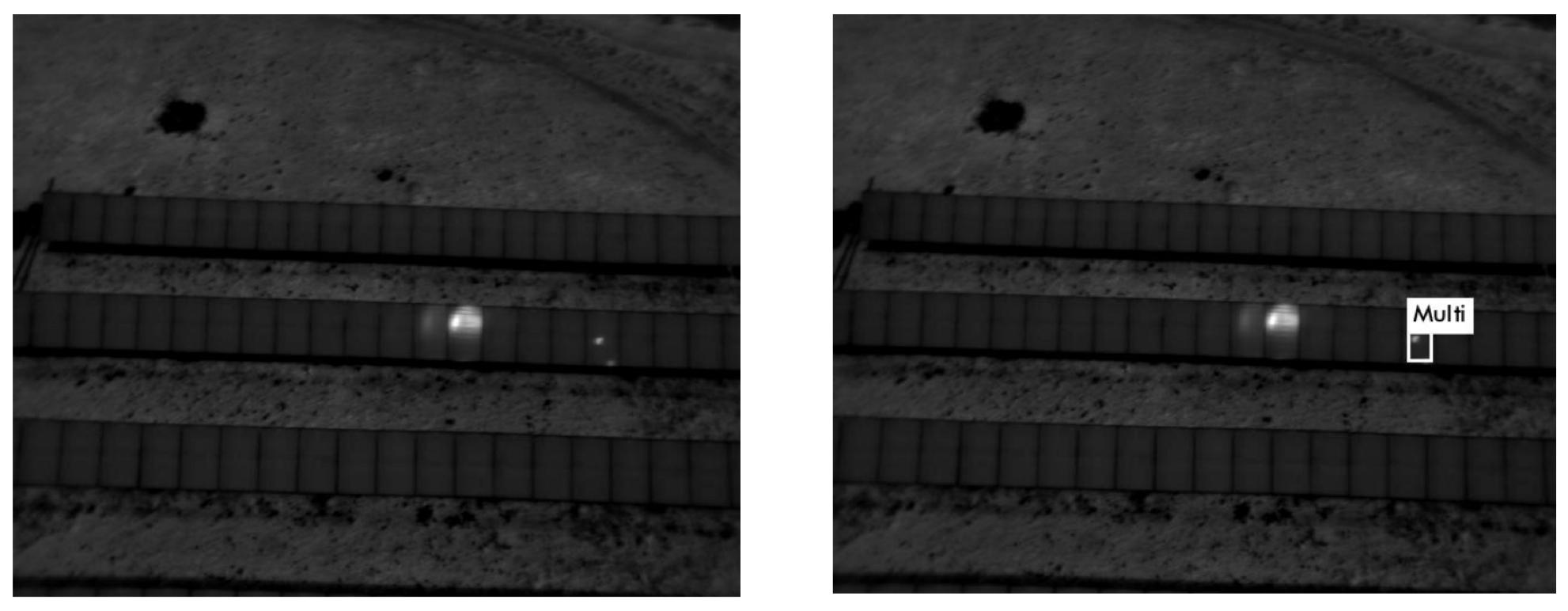}
  \caption{Model outputs for multi cell hotspots. Notice that the model ignores sun's glare in the second pair.}
  \label{mtyp_defects}
\end{figure}

\begin{figure}
  \centering
  
  \includegraphics[width=0.6\paperwidth]{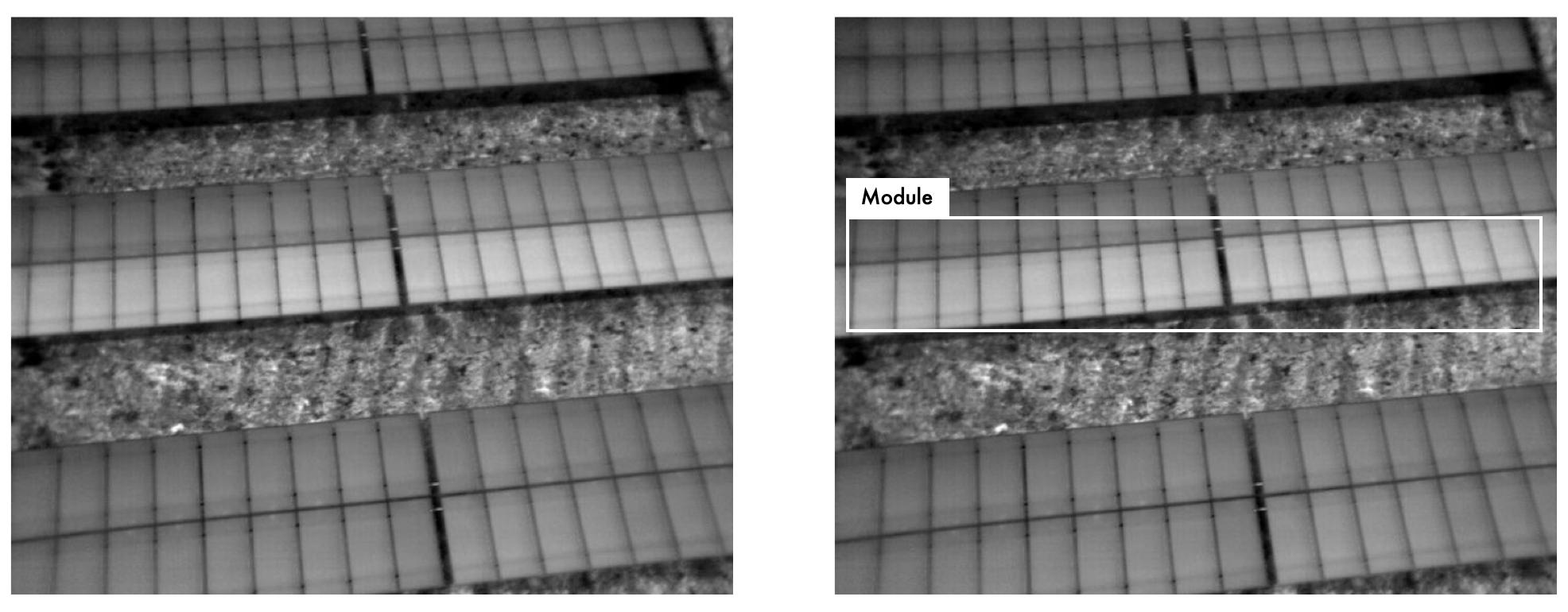}
  \caption{Model output for module hotspots.}
  \label{modtyp_defects}
\end{figure}
\subsection{Observations}

We see that the model learns to identify the hotspots and distinguish between the classes. The bounding box generated is fairly tight which helps to localise the hotspot over specific panels. We observe that the model has learnt to ignore sun's glare which is also seen as a bright hotspot as shown by figure 3.

Figure 2,3,4 show the results of detection of single cell hotspots, multi cell hotspots and module hotspots repectively.

\section{Conclusion}

We show how deep learning based object detection can be used to detect defects in solar panels. This is a time and cost efficient way to imporve the output of solar power plants. We have scanned more than 150MW of solar power plants around India and once repaired, the energy saved can power more than 100 urban homes.

\medskip

\small

\bibliographystyle{unsrt}
% \bibliography{biblio}

\end{document}